%% file: ms.tex
\documentclass[conference, svgnames]{IEEEtran}

\usepackage{amssymb}
\usepackage{amsmath}
\usepackage{amsfonts}

\usepackage[ruled,vlined,linesnumbered]{algorithm2e}
\SetCommentSty{mycommfont}

\usepackage{booktabs} % For formal tables
\usepackage{array}
\usepackage{graphicx}
\usepackage{todonotes}
\usepackage{romannum}

\usepackage{nicefrac}
\usepackage{bm}

\usepackage{booktabs} % for professional tables

\usepackage{caption}
\usepackage{subcaption}
\usepackage{multicol}
\usepackage{multirow}

\usepackage{float}

\newcommand{\ignore}[1]{}

\usepackage{listings}
\usepackage{pifont}

\newcommand{\eg}{\textit{e.g.}\xspace}
\newcommand{\ie}{\textit{i.e.}\xspace}

\newcommand{\squishitemize}{
 \begin{list}{$\bullet$}
  { \setlength{\itemsep}{-0pt}
     \setlength{\parsep}{2pt}
     \setlength{\topsep}{0pt}
     \setlength{\partopsep}{0pt}
     \setlength{\leftmargin}{1.0em}
     \setlength{\labelwidth}{1.0em}
     \setlength{\labelsep}{0.5em} } }

\newcounter{Lcount}
\newcommand{\squishlist}{
    \begin{list}{\arabic{Lcount}. }
   { \usecounter{Lcount}
        \setlength{\itemsep}{-0pt}
        \setlength{\parsep}{3pt}
        \setlength{\topsep}{0pt}
        \setlength{\partopsep}{0pt}
        \setlength{\leftmargin}{2em}
        \setlength{\labelwidth}{1.5em}
        \setlength{\labelsep}{0.5em} } }

\newcommand{\squishend}{\end{list}}

\usepackage{gensymb} % for \degree

\usepackage{cite}

\definecolor{Mycolor2}{HTML}{007FFF}
\newcommand\hl[1]{%
  \bgroup
  \hskip0pt\color{Mycolor2}%
  #1%
  \egroup
}

\definecolor{linkcolors}{HTML}{CC0000}

\usepackage[hyphens]{url}
\usepackage[colorlinks, pagebackref=false, breaklinks=true, letterpaper=true,bookmarks=true, linkcolor=black, citecolor=black, urlcolor=black]{hyperref}

\usepackage[capitalise,noabbrev]{cleveref}
\crefformat{section}{Section #2#1#3} % see manual of cleveref, section 8.2.1
\crefformat{subsection}{Section #2#1#3}
\crefformat{subsubsection}{Section #2#1#3}

\begin{document}

\title{\vspace{-20pt} Context-Aware Task Handling in Resource-Constrained Robots with Virtualization
}

\author{
  Ramyad Hadidi, Nima Shoghi Ghalehshahi, Bahar Asgari, Hyesoon Kim\\
  Georgia Institute of Technology\\
}

\maketitle

%%%%%%%%%%%%%%%%%%%%%%%%%%%%%%%%%%%%%%%%%%%%%%%%%%%%%%%%%%%%%%%%%%%%%%%%%%%%%%%%
\begin{abstract}
\input{tex/abs.tex}

\end{abstract}


%%%%%%%%%%%%%%%%%%%%%%%%%%%%%%%%%%%%%%%%%%%%%%%%%%%%%%%%%%%%%%%%%%%%%%%%%%%%%%%%
\section{Introduction \& Motivation}
\label{sec:intro}
\input{tex/intro.tex}

% %------------------------------------------
\section{Related Work}
\label{sec:related}
\input{tex/related.tex}
\section{Deconstructing Tasks}
\label{sec:tasks}
\input{tex/tasks.tex}

% %------------------------------------------
\section{Context-Aware Task Handling}
\label{sec:technique}
\input{tex/tech.tex}

% %------------------------------------------
\section{Module Implementations}
\label{sec:task-description}
\input{tex/tesk-description}

\section{Experiments}
\label{sec:res}
\input{tex/res.tex}

% %------------------------------------------
\section{CONCLUSION}
\label{sec:conclusion}
\input{tex/conclusion.tex}

%%%%%%%%%%%%%%%%%%%%%%%%%%%%%%%%%%%%%%%%%%%%%%%%%%%%%%%%%%%%%%%%%%%%%%%%%%%%%%%

%%%%%%%%%%%%%%%%%%%%%%%%%%%%%%%%%%%%%%%%%%%%%%%%%%%%%%%%%%%%%%%%%%%%%%%%%%%%%%%%
%\section*{APPENDIX}
%
%\section*{ACKNOWLEDGMENT}

%%%%%%%%%%%%%%%%%%%%%%%%%%%%%%%%%%%%%%%%%%%%%%%%%%%%%%%%%%%%%%%%%%%%%%%%%%%%%%%%

\bibliographystyle{IEEEtran}
\bibliography{extracted}

\end{document}

%% file: tex/abs.tex
%
%
%
% Mobile robots are critical in countless scenarios such as swarm and collaborative tasks. Building intelligent mobile robots include the integration of several tasks such as manipulation, navigation, planning, reasoning, and sensing.  As computational resources are limited in mobile robots, they struggle in handling several tasks concurrently and yet guaranteeing real-timeliness. Nevertheless, we observe that according to the context, only a few tasks are critical at a moment, hence not all the tasks need to be executed concurrently. As a result, to extend the range of tasks for mobile robots and to improve the real-timeliness of critical tasks under resource restrictions, our key insight is to efficiently reduce the number of tasks at each moment to only the critical ones. To do so, we propose a \textit{fast context-aware} task handling technique. To effectively handling tasks in \textit{real-time}, our proposed context-aware technique comprises of three main ingredients: (i) a dynamic time-sharing mechanism, coupled with (ii) an event-driven task scheduling using reactive programming paradigm to mindfully use the limited resources; and, (iii) a lightweight virtualized execution to integrate several functionalities with different dependencies while enabling a kernel-level dynamic task scheduling policy. We showcase our technique on a Raspberry-Pi-based robot with a variety of tasks such as SLAM, sign detection, and speech recognition with a 42\% speedup in total execution time compared to the common Linux scheduler.
%
%
Intelligent mobile robots are critical in several scenarios. However, as their computational resources are limited, mobile robots struggle to handle several tasks concurrently and yet guaranteeing real-timeliness. To address this challenge and improve the real-timeliness of critical tasks under resource constraints, we propose a \textit{fast context-aware} task handling technique. To effectively handling tasks in \textit{real-time}, our proposed context-aware technique comprises of three main ingredients: (i) a dynamic time-sharing mechanism, coupled with (ii) an event-driven task scheduling using reactive programming paradigm to mindfully use the limited resources; and, (iii) a lightweight virtualized execution to easily integrate functionalities and their dependencies. We showcase our technique on a Raspberry-Pi-based robot with a variety of tasks such as Simultaneous localization and mapping (SLAM), sign detection, and speech recognition with a 42\% speedup in total execution time compared to the common Linux scheduler.

%% file: tex/intro.tex
%
%
 
% While conventional industrial or commercialized robots perform a set of pre-programmed and routine tasks, intelligent mobile robots are designed to navigate ~\cite{liu2014navigation, liu2013visual}, plan~\cite{liu2015robotic, colas20133d}, reason, recognize~\cite{liu2012dp}, and sense~\cite{liu2015incremental, liu2013information} accordingly that matches their environment and goals.
%
%
%%%% SEVERAL TASKS TO RUN: 
Unlike conventional industrial or commercialized robots that perform a set of pre-programmed and routine tasks, intelligent mobile robots manipulate their environment using their perception and physical resources to achieve a myriad of goals. Such robots must be capable of dynamically switching between navigation, planing, reasoning, recognition, and sensing their environment. Intelligent robots need to interact with a dynamic, complex, and non-deterministic world. These robot must execute numerous tasks such as controlling its physical resources (\eg, arms), processing and understanding data derived from sensors, or executing perception and planning.

%%%% KEY CHALLENGE OF LIMITED RESOURCES:
Intelligent robots are always in a never-ending conflict between available computation resources, their energy storage, and the tasks at hand. This conflict is particularly emphasized in resource-constrained robots because even the concurrent execution of a few rudimentary tasks is extremely demanding with only a few processing cores. For example, a Raspberry Pi with only four cores, could be fully utilized by the operation system (OS), processing the data from a single sensor, and simple navigation and control algorithms. Adding more sensors and tasks only causes the robot to miss real-time deadlines. Thus, ensuring efficient handling of critical tasks and meeting the critical deadlines is the key challenge for resource-constrained robots.

% Moreover, executing complex solver-style planning algorithms to ensure appropriate planning is costly both in computational power and losing processor time. 

%%%% AVAILABLE SOLUTIONS (PRIOR WORK):
To extend the capabilities of resource-constrained robots and meet real-time demands, the common practices are adding extra hardware or utilizing cloud/fog computation~\cite{kehoe2015survey, hadidi2018distributed, wang2016hierarchical}. However, in several scenarios, adding new hardware are either infeasible or uneconomical. For example, adding extra processing units to a lightweight drone requires heavier batteries, which in turn demands stronger motors. Further, cloud and fog are not always available. Additionally, privacy concerns limits the suitability of cloud-based computation.

% guizzo2011robots, wang2015real, kattepur2017priori
% incur communication overheads, have unreliable connections,

%%%% KEY INSIGHT:
To enable intelligent mobile robots to efficiently utilize limited resources, we propose a \textit{context-aware} task handling technique that simplifies the world and planning tasks by dynamically reducing the number of tasks in a certain context to only the critical ones. For example, limited human-robot interaction is expected while the robot is performing an already assigned task. This technique enables resource-constrained robots to efficiently perform manifold functionalities while meeting their real-time constraints.

% For instance, in sign reading, instead of executing the entire recognition task continuously, the robot can execute lighter bounding box detection task first. Then, if the presence of a sign is likely, the robot may try to execute the entire task to understand the sign. Moreover, a user can input a configuration file that defines a custom context-aware setting.

%%%% OTHER CONTRIBUTIONS:
To be effective in handling tasks using our context-aware technique, we propose using a \textit{virtualized execution} that (i) integrates several tasks while providing dynamic, low-cost, and kernel-level control over the scheduling policy; (ii) enables easier context-aware implementation by providing manageable control over tasks; and (iii) provides a uniform and practical environment for building new robots in the community.

% with the ubiquity of single-board computers (\eg, Raspberry Pi) and robots based on them. %

% We target resource-constrained and inexpensive robots since (i) they are currently only limited to rudimentary tasks because of their limited computational resources, and (ii) they are considered the main option for several tasks and environments (e.g., light drones for surveying an area after a disaster).

\begin{figure}[t]
  \vspace{-10pt}
  \centering
  \includegraphics[width=0.80\linewidth]{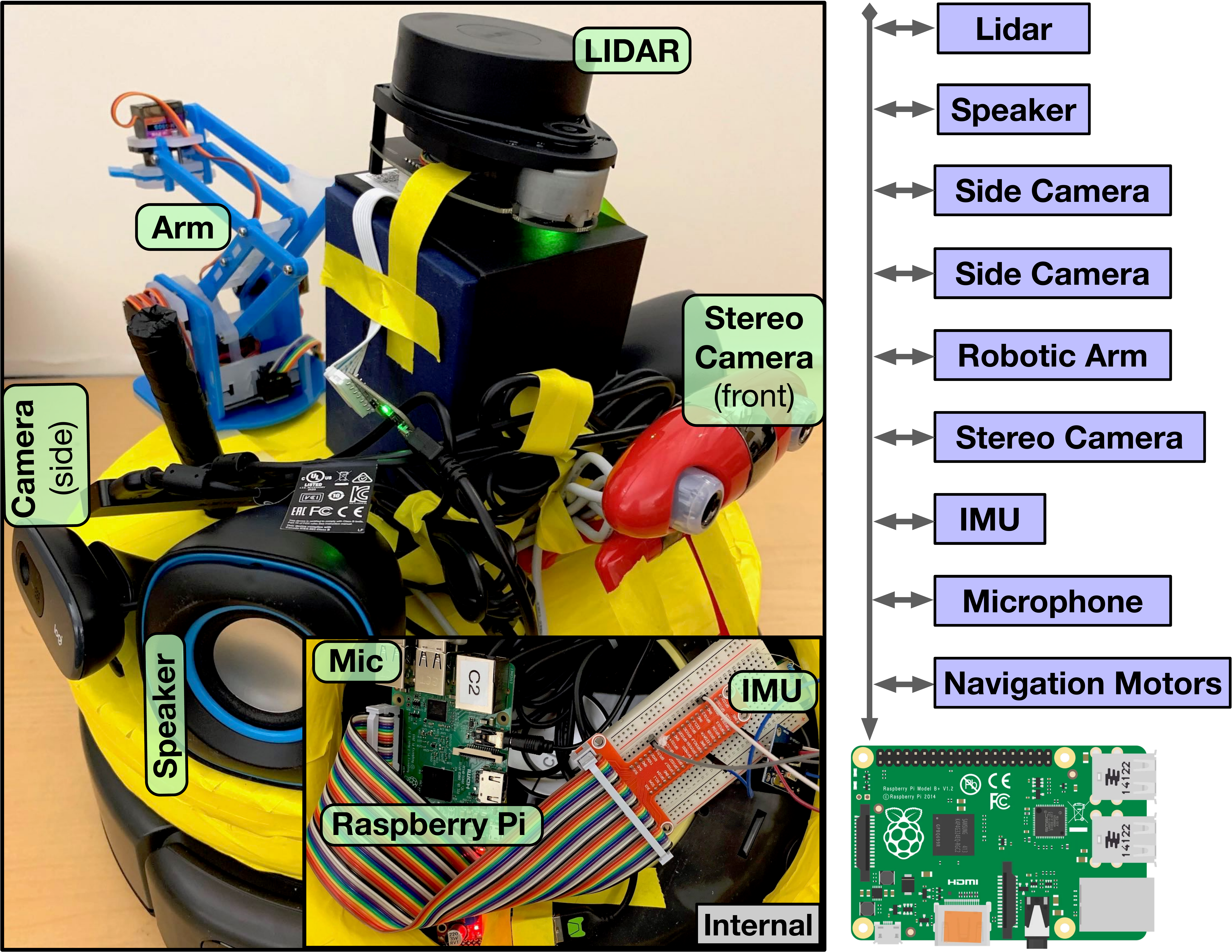}
  \vspace{-2pt}
  \caption{\small Modified iRobot with RPi4 and additional sensors.}
  \vspace{-15pt}
  \label{fig:irobot}
\end{figure}

%%%% IMPLEMENTATION:
For experiments, we use a custom-built Raspberry-Pi-based robot using an iRobot Roomba~\cite{irobot} equipped with one Raspberry Pi 4 (RPi4)~\cite{pi4} as the only processing unit. Our iRobot, shown in Figure~\ref{fig:irobot}, has several sensors (\ie, LIDAR, inertial measurement unit (IMU), cameras, and microphone), and control devices (\ie, motors for navigation, robotic arm, and speakers). For software, we use Docker~\cite{docker}, a popular virtualization tool and implement our context-aware technique to collect and process sensors data, simultaneous localization and mapping (SLAM), voice recognition, and sign recognition.
Our contributions are as follows:
\begin{itemize}
    \item \emph{Context-aware} task planning to effectively use the limited resources and hence extend the number of tasks that a robot can handle.
    \item OS-level \emph{dynamic} time-sharing to implement the context-aware scheduling in real-time.
    \item \emph{Event-driven} task scheduling to be mindful in using the limited resources for scheduling itself.
    \item \emph{Lightweight virtualized execution}, using Docker and reactive programming paradigm to enable easily manageable and yet kernel-level dynamic task scheduling policy.
\end{itemize}

% Furthermore, we emulate complex real-world scenarios by mixing common datasets for each task.

% Finally, we have released our framework and experiments as supplementary materials.

%
%

%% file: tex/related.tex
\subsubsection*{Real-Time Operating Systems \& Scheduling Polices}
%
% In 90s, real-time performance had been studies for time-sharing in general-purpose operating systems with approaches such as reserving a certain amount of CPU time~\cite{lee1996experiences, jones1997cpu}, and using LXRT module in RTAI/Linux (LXRT) or utilizing fixed priorities defined within POSIX4 to permit hard and soft real-time tasks in user-space context, respectively. Later, real-time operating system (RTOS) has been designed to process data in real-time aiming to minimize the latency of accepting a process instead of targeting for higher throughput in conventional operating systems (OSs)~\cite{laplante1996real, tanenbaum2015modern}.

The operating system (OS) schedules applications either based on the order of the events (event-driven), order of processes (\eg, round-robin), or time sharing. To minimize the latency of accepting a process real-time operating system (RTOS) has been designed. RTOSes have preemptive schedulers~\cite{ramamritham1994scheduling} (\eg, fixed-priority preemptive scheduling). Since optimal scheduling is an NP-complete problem~\cite{ramamritham1994scheduling, tanenbaum2015modern}, even RTOSes can not guarantee hard deadlines. Therefore, hard real-time robotic systems usually either implement fixed schedulers (\eg, commercial drones) or use extra dedicated cores to provide enough computation performance. As neither solutions aligns with our goal of \textit{context-aware} task handling using \textit{limited resources}, this paper tunes the OS scheduler (\cref{sec:technique:policy}).

% which indicates that the OS has a clock interrupt to wake the scheduler so that it can switch to a higher priority task

%-------------------------------------------------------------------
% \subsubsection{Environments for Robotics}
%
% \subsubsection*{Robotic Development Environments (RDEs)}
\subsubsection*{Robot Operating System (ROS)}
 Robot Operating System (ROS)~\cite{quigley2009ros} is a popular example of robotic environments to manage the complexity of various aspects of robotic systems, from simulation to hardware implementation. ROS also manages the process execution, while providing stand-alone libraries for hardware components. As ROS does not offer real-time operations, ROS2 has been upgraded to handle hard real-time tasks~\cite{casini2019response} by prioritizing real-time threads and avoiding the sources of non-determinism such as memory allocation~\cite{kay2015real}. Nevertheless, ROS2 does not support dynamically changing priorities in runtime. Moreover, ROS2 requires additional kernel support~\cite{rt-patch}, still in early development.

%% file: tex/tasks.tex
%
%
% \subsection{Task Categories}
% \label{sec:task:theory}
%
% An intelligent robot performs various tasks, each of which comprises several parts. We use a variant of the terminologies presented by \cite{zlot2006auction} for describing the type of tasks.

To design our context-aware task handling, we first categorize tasks as the following: The first category is \emph{elemental} or atomic tasks that consist of a single event. The second category, \emph{compound} task, is \emph{decomposed} into multiple steps or a set of subtasks. To satisfy a \emph{compound} task, every sub-task of it must be done. The third category, \emph{complex} tasks, are also decomposable into subtasks, whereas to satisfy a complex task, not all subtasks are required to be done.

% Thus, the main difference between compound and complex tasks is the possibility of different ways of satisfaction. For instance, for task $t$ that can be decomposed to a total of $n$ $\varepsilon$ sub-tasks, or $t=\{\varepsilon_1, \varepsilon_2, ..., \varepsilon_n \}$, if to satisfy $t$, we need to satisfy all of the $n$ sub-tasks, then $t$ is a \emph{compound} task. Otherwise, if for satisfying $t$, we can satisfy certain sets of possible combinations of $\varepsilon$, then $t$ is a complex task. In other words,
% %
% %
% \begin{equation}
% \begin{split}
%     \small
%     \text{For task $t = \{\varepsilon_i\mid i \in [1,n]\}$, to satisfy $t$}
%     \\
%     \text{}
%     \begin{cases}
%         \small
%         \text{do all $\varepsilon_i \mid i \in [1,n]$} & \text{\small if $t$ is compound,}\\
%         \text{do a set $\{\varepsilon_i\}\mid i \in [1,n]$} & \text{\small if $t$ is complex.}
%     \end{cases}
% \end{split}
% \vspace{-15pt}
% \end{equation}
%
%

%
\begin{figure}[t]
  \vspace{-10pt}
  \centering
  \includegraphics[width=0.75\linewidth]{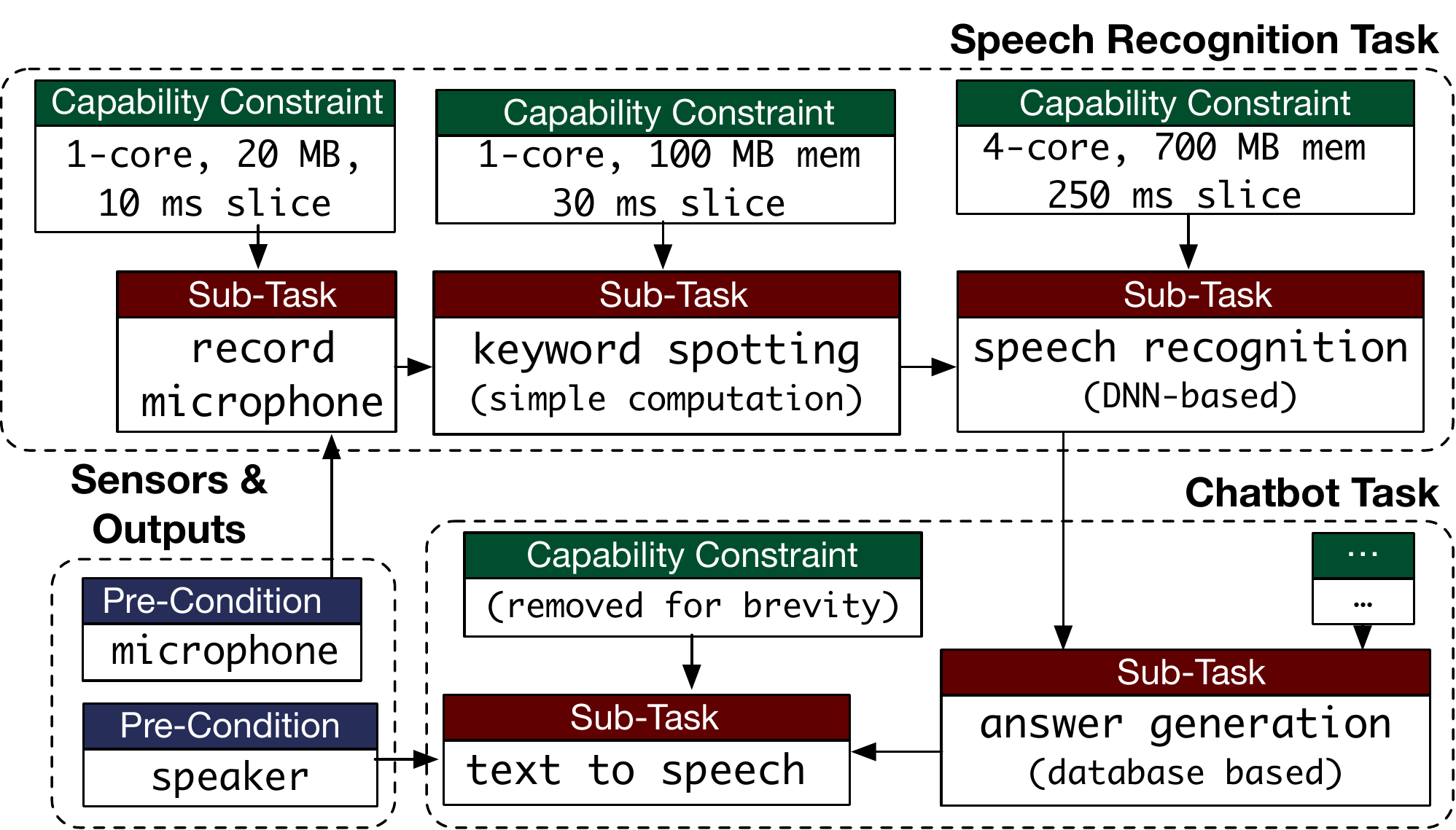}
  \vspace{-2pt}
  \caption{\small Graph representation for speech recognition and chatbot tasks. A representation of $(\{{\varepsilon_{t}}_i\}_{i=1:n}, \rho_t)$.}
  \vspace{-15pt}
  \label{fig:task-graph-example}
\end{figure}
%

%-------------------------------------------------
% \subsection{Relationships of Sub-tasks, Constraints, \& Conditions}
% \label{sec:task:const}
%
Resulting subtasks have a set of relationships with each other, possible pre-conditions, and capability constraints. For instance, the \emph{pre-condition} of executing the speech recognition task is to have a speech input. In this case, the speech recognition task has a \emph{relationship} with speech input. Besides, tasks have relationships with the \emph{capability constraint} to execute a workload within a deadline. For instance, to execute speech recognition effectively, we require full access to all the cores of the processor and a certain amount of memory. For a task $t$, we show such relationships with a directed graph structure, $\rho$, the vertices of which are sub-tasks/conditions/constraints and its edges are the relationships. Therefore, the pair $(\{{\varepsilon_{t}}_i\}_{i=1:n}, \rho_t)$, for a task $t$, represents all sub-tasks, conditions, constraints, and relationships. Figure~\ref{fig:task-graph-example} illustrates an example for speech recognition and chatbot tasks. For instance, keyword spotting processes microphone recording and requires a single-core and 100\,MB memory. A compiler analysis can extract this graph automatically. In the following, we present a manual low-overhead approach.

% In this example, speech recognition has three sub-tasks, recording from a microphone, keyword spotting, and speech recognition. Each sub-task has some relationship with other sub-tasks, capability constraints, or pre-conditions.

% Similarly, the chatbot task has a relationship with the speech recognition task and requires the availability of the speaker.

%--------------------------------------------------------------
\subsubsection*{Containerizing Modules}
\label{sec:technique:docker}

In the first step, each independent task is wrapped as containerized modules implemented as Docker~\cite{docker} containers. Docker implementations are easy to configure and distribute. Meanwhile, since the lower-level OS abstraction and common libraries and dependencies are shared, the overhead of using Docker is minimal.

%--------------------------------------------------------------
\subsubsection*{Adding Event-Driven Initiatives}
\label{sec:technique:rx}

By using Reactive Extensions (RX) framework~\cite{reactivex}, next, the user adds a simple event-driven initiative for each module. This declarative configuration sets \emph{scheduling scores} (more in~\cref{sec:technique:controller}) of modules while abstracting low-level implementation (\eg, synchronization, thread-safety, concurrent data structures, and non-blocking I/O). RX provides tools for operating on, filtering, and managing asynchronous streams of data. Such streams are called \emph{observable streams} and indicate sensor readings over time.  For example, in the below example, the inertial measurement unit (IMU) sensor is a single observable stream of accelerometer and gyroscope readings over time. We filter this stream for receiving readings that have a non-zero vector.
\begin{lstlisting}[language=Python, linewidth = 1.0\linewidth, caption={Constructing observable streams for IMU.}] 
imu.pipe(filter(lambda value:
    value["accelerometer"]["x"] != 0 and
    value["accelerometer"]["y"] != 0 and
    value["accelerometer"]["z"] != 0))
   
\end{lstlisting}
\vspace{-5pt}
%
%

% inspired by the functional reactive programming paradigm~\cite{hudak2002arrows}

% to effectively create an event-driven system based on dynamic inputs. RX extends the observer design pattern~\cite{hannemann2002design} for supporting asynchronous events.

% To manage observable streams, RX provides a set of operators and utility functions.

% \footnote{The \texttt{imu} variable represents the observable stream corresponding to the IMU sensor reading}

% by using \texttt{filter} operator

%% file: tex/tech.tex
\begin{figure}[t]
  \vspace{-10pt}
  \centering
  \includegraphics[width=0.80\linewidth]{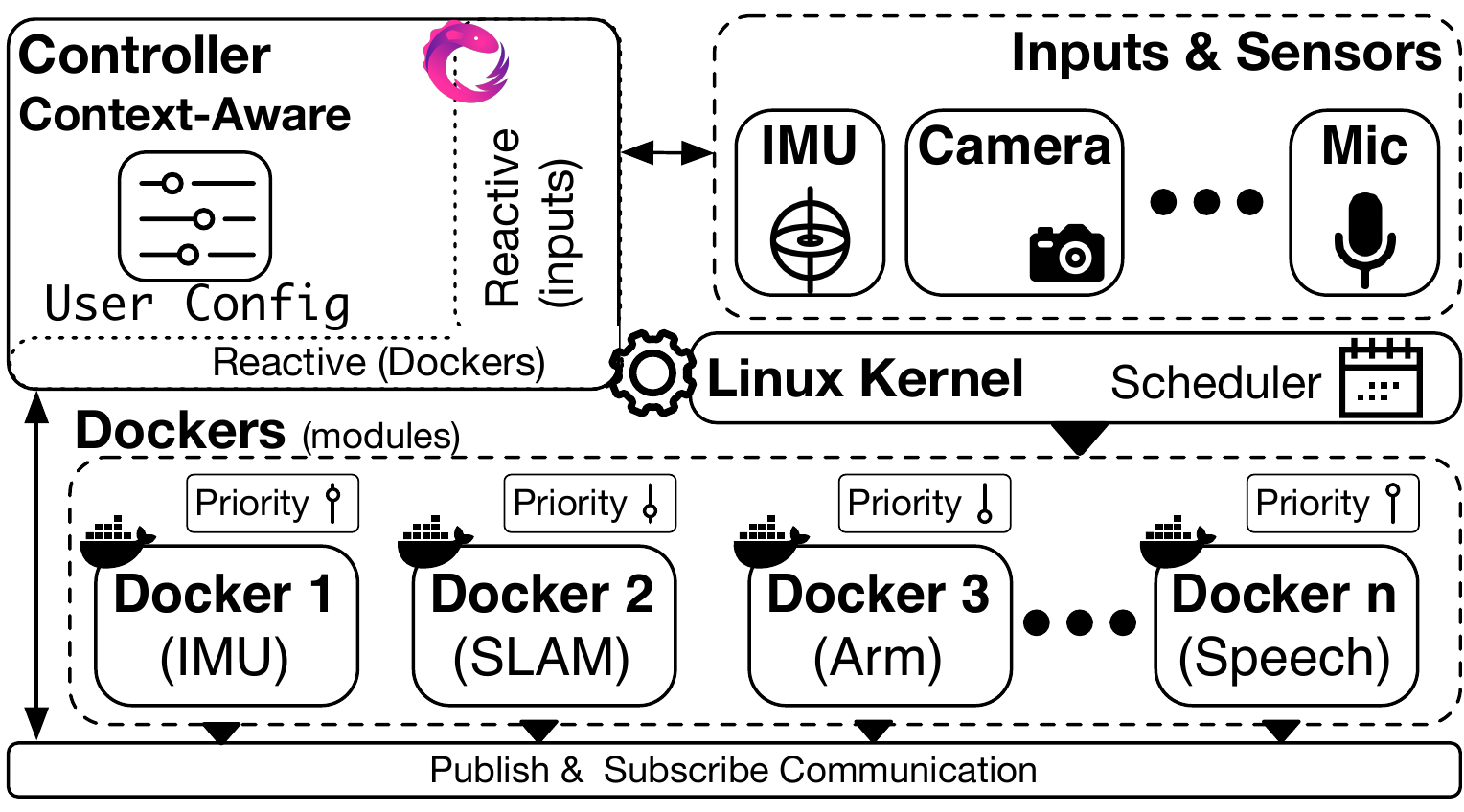}
  \vspace{-2pt}
  \caption{\small High-level system overview.}
  \vspace{-10pt}
  \label{fig:system-overview}
\end{figure}

To reduce the number of tasks at each moment to only the critical ones, we propose a \textit{context-aware} and \textit{event-driven} task handling technique, a high-level overview of which is shown in Figure~\ref{fig:system-overview}. The implementation is separated into two groups: (i) the \emph{controller}, and publish and subscribe communication medium, which manages the dynamic OS-level prioritizing/scheduling and communications among modules, respectively, and (ii) the modules that carry out the tasks, represented as containers, or \emph{Dockers}. This section describes the first group, and how they achieve mentioned goals.

% We first describe the publish-subscribe communication (\cref{sec:technique:publish}). Then, we explain the controller (\cref{sec:technique:controller}), scheduling policy (\cref{sec:technique:policy} and \cref{sec:technique:scores}), and, finally, integrating all components in task planning procedure (\cref{sec:technique:all}).
 
%  The controller is a lightweight program that controls and configures how entire system behaves by configuring OS-level scheduling policies.

%
%

%--------------------------------------------------------------
\subsection{Publish-Subscribe Communication}
\label{sec:technique:publish}
Publish-subscribe communication pattern provides an efficient medium for sending and receiving data. While the modules are usually isolated and operate independently of each other, communication is necessary (\eg, when the navigation requires mapping information from the SLAM). We implement a lightweight and resource-efficient publish-subscribe system among modules, in which the events can subsequently be wrapped with an RX observable stream and fed back into the controller. This enables to input any output of a module to the controller if needed. Additionally, the shared memory interface is used to share parsed binary information among applications. This is especially efficient if such information is already serialized without the extra cost of repacking.

% The inter-process communication (IPC) interface of POSIX in our underlying implementation uses message queues for efficiently setting up a bi-directional event passing system,

% For instance, after a ``left'' command, we can deprioritize the command detection module for a short period since the robot is performing the task.

% (serialization mechanisms such as FlatBuffers~\cite{oortmerssen2014flatbuffers} or Cap'n Proto~\cite{varda2015cap}).

%--------------------------------------------------------------
\subsection{The Controller}
\label{sec:technique:controller}
The controller is a lightweight program for dynamically setting the priorities of modules based on the context. Based on added event-driven initiatives (\cref{sec:tasks}), the controller uses incoming sensors and modules data to dynamically decide the best scheduling scores or weights, $w$. To calculate scheduling scores, the controller processes the observable streams and if it detects a context change, it will change the priorities accordingly in the Linux kernel scheduler. The calculation of scheduling scores executes only when the dependent sensor/module values update, saving valuable calculation time on resource-constrained platforms.

\begin{figure}[t]
  \vspace{-10pt}
  \centering
  \includegraphics[width=1.0\linewidth]{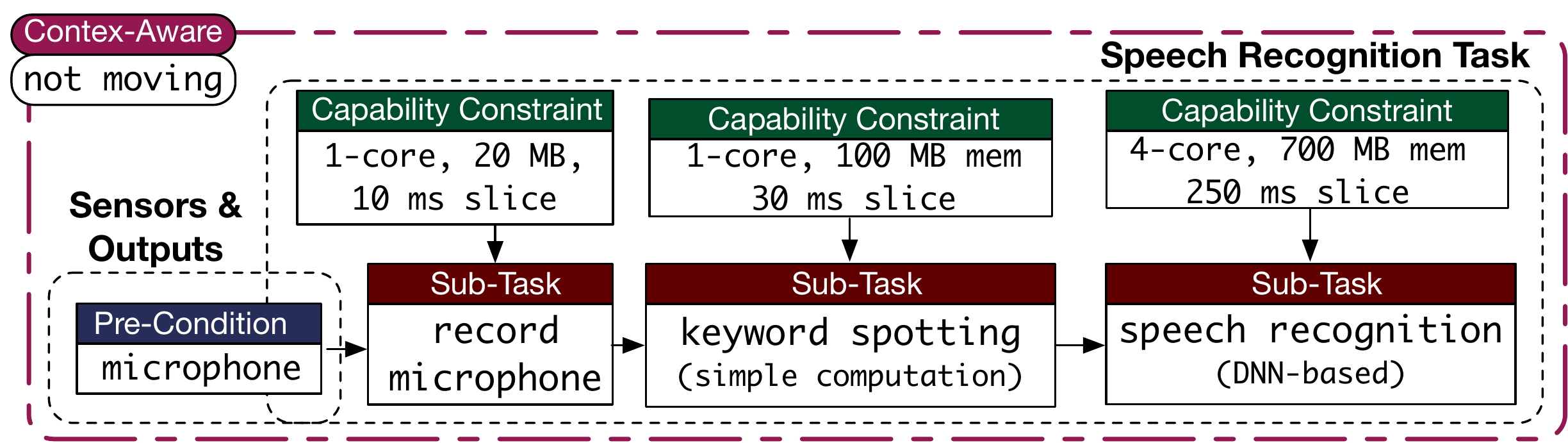}
  \vspace{-13pt}
  \caption{\small The modified context-aware graph representation for speech recognition with a simple context-aware condition.}
  \vspace{-10pt}
  \label{fig:task-graph-example-context-aware}
\end{figure}

To give users power to define contexts, the controller can also receive an input from the user with a configuration file. This configuration file first specifies the relative priority of the tasks, and second, it may extend the context-aware decisions in the controller. For instance, the user may specify that no microphone-related task should run while the robot is moving. Therefore, the speech recognition task never executes while the robot is moving, and accordingly, no microphone input is processed. In other words, the controller dynamically modifies the task graphs based on sensor events and user configuration, which leads the system to automatically adapt new scheduling scores. An example is shown in Figure~\ref{fig:task-graph-example-context-aware} by modifying speech recognition task in Figure~\ref{fig:task-graph-example}.

% based on which scheduling parameters are calculated (details in \cref{sec:technique:scores}).

% Then, the controller calculates the scheduling scores for all the modules and sets each prioritization accordingly in the Linux kernel scheduler. 

% This is realized by using a configuration for each module that includes a simple function to subscribe to any of the observable streams. If necessary, these streams in a module can be combined to first generate a floating-point value, and then, if necessary, wake up the controller. Thus, the controller only calculates the weights when is necessary, creating context-aware task graphs. 

%
%

%--------------------------------------------------------------
\subsection{Scheduling Policy}
\label{sec:technique:policy}
In robotics, real-time functionality and scheduling customizability are important. However, by default, Linux uses the completely fair scheduler (CFS, \texttt{SCHED\_OTHER})~\cite{sched-linux-manual} to provide an optimal setting for desktops and servers. As reviewed in \cref{sec:related}, ROS2 requires extra kernel support for real-time prioritization and still is unstable. To address these issues, while relying on stable and fully-supported Linux features, we dynamically tune the parameters of the scheduler per module as we receive real-time value updates. We apply this approach with two methods: (i) using CFS policies (\ie, \texttt{SCHED\_OTHER}) and tuning the CFS parameters of each module, and (ii) Using real-time policies~\cite{sched-linux-manual} (\ie, \texttt{SCHED\_FIFO} or \texttt{SCHED\_RR}) and adjusting real-time parameters of each module. The following provides the implementation details.

% To mitigate these issues, we propose a general solution: We tune the parameters of the scheduler used for the module as we receive real-time sensor information. 

% However, current versions of Linux provide other scheduling algorithms and configurations, such as the multiple queue skiplist scheduler (MuQSS)~\cite{kolivas2016muqss} and the real-time scheduling policies. 

% CFS provides an optimal experience for desktop and server use cases, but it is not optimal for robotics, where real-time functionality and scheduling customizability is important. 

% pabla2009completely

\subsubsection{Tuning the CFS Parameters}
\label{sec:technique:policy:cfs}
In CFS, we use the \texttt{cpu-period} and \texttt{cpu-quota} flags in modules to customize resource allocation. \texttt{cpu-quota} is the total amount of CPU time that a module can use in each \texttt{cpu-period}. For this feature, note that Linux kernel should be compiled with CFS bandwidth control flag~\cite{sched-bwc}. Additionally, Docker provides a combined flag, \texttt{cpus}, which allows us to directly allocate CPU resources to a container.

% without having to run the tasks with a real-time policy option.

% is allowed to use, represented as a fraction of the total number of physical and virtual processors in the system.

\subsubsection{Adjusting Real-Time Parameters}
\label{sec:technique:policy:rt}
Although Linux real-time (RT) policies~\cite{sched-linux-manual} provide a better determinism to processes, the policies do not allow changes to the priorities during runtime. In details, in both \texttt{SCHED\_FIFO} or \texttt{SCHED\_RR} policies, each process gets a time-slice or exclusive access to the CPU defined during the process startup (\texttt{SCHED\_FIFO} runs real-time processes until it finishes. \texttt{SCHED\_RR} builds on top of \texttt{SCHED\_FIFO} by implementing a round-robin time-slice system based on some priority). To tune real-time scheduling parameters during runtime, we limit the the total number of microseconds each module runs using Docker at real-time priority by setting the \texttt{cpu-rt-runtime} with the controller. This flag is set to a value between 0$\mu$s and 1s, and it represents the total number of microseconds reserved. We use this feature to use contextual information to change the real-time resource allocation for each module dynamically. In summary, using Docker and RX, we are able to create a dynamic two-level scheduler that is (i) event-based due to reactive programming paradigm of the controller, (ii) time-sharing due to the Docker ability for setting time-slice value (in our case, through the controller), and (iii) dynamic because the controller changes the time-slices during runtime based on the context (supplied by user and extracted from tasks).

% However, it is important to note that the processes (\eg, sign recognition) do not need to be ran in real-time mode. We expect run-time speedups even if these processes are scheduled with the normal \texttt{SCHED\_OTHER} policy.

% This is the reason why our controller (\cref{sec:technique:controller}) tunes scheduling parameters during runtime.

% are assigned time-slices based on their priority and run until the end of their time-slice.

%  Therefore, the user can choose between an event-based and a time-sharing policy. Moreover, the priority of the time-sharing policy is fixed and defined during the process startup. Therefore, by only using these policies, . 

% For this feature, the kernel should be compiled with configuration real-time group scheduling flag~\cite{sched-rt-group}. 

% 

% Running processes using real-time policies must be done with caution, as this indicates that the operating system will prioritize these processes even over kernel processes.

%-------------------------------------------------------
\subsection{Calculating Scheduling Parameters}
\label{sec:technique:scores}

Here, we describe how the controller calculates scheduling parameters for each module based on the context. Every module defines (automatically or from user) an instantiation function that returns an observable stream representing the scheduling score as a floating-point value, or $w_{i}$. This function can take any RX stream as its input (from sensors or other modules). When the system starts, the controller calls all the instantiation functions and creates observable streams which produces floating-point values representing the scheduling score. Then, the controller combines all of the observable streams into a single observable stream. This aggregated stream is an observable stream that output the entire set of scheduling scores. For each container $c$ with scheduling score $w_{c}$ (specified by user and/or from context) received from the aggregated observable stream, we calculate the scheduling parameters for two scheduling types (\cref{sec:technique:policy}) as follows. 
For CFS (\cref{sec:technique:policy:cfs}), the CPU share value (\texttt{cpus}), $s_{c}$, is calculated with the equation below, where $N$ is the number of processors in the system:
\begin{equation}
  \small
  s_{c} \\= N \cdot \frac{w_{c}}{\sum_{i}{w_{i}}}. 
  \label{equ:final_score_cfs}
\end{equation}
For real-time scheduler (\cref{sec:technique:policy:rt}), the real-time time-slice value (\texttt{cpu-rt-runtime}), $t_{c}$, is calculated as below, where $P$ is the time-slice period, which is 1s by default:
\begin{equation}
  \small
  t_{c} \\=  P \cdot \frac{w_{c}}{\sum_{i}{w_{i}}}. 
  \label{equ:final_score_rt}
\end{equation}
The result of the above expression creates a dictionary that maps each container, $c$, to its CPU share value, $s_{c}$ or real-time time-slice value, $t_{c}$. Note that the aggregated stream is also an event-driven operation with all the RX capabilities in filtering not-related events. Thus, the controller only updates the time-slices if it observes any new event.

\input{tex/algo.tex}

% The controller subscribes to this stream and sets the \texttt{cpus} or \texttt{cpu-rt-runtime} values of each container as it calculates new values over time.  Figure~\ref{fig:timelines} shows a simple system with only two observable streams and their calculated scores. The aggregated stream uses \texttt{combineLatest} to combine the events and calculates new time-slices values.

% \begin{figure}[b]
%   \vspace{-15pt}
%   \centering
%   \includegraphics[width=0.9\linewidth]{fig/timelines.pdf}
%   \vspace{-0pt}
%   \caption{\small Aggregated stream with two observable stream for SLAM and audio. The aggregated stream combines scheduling scores and the controller calculates the time-slice value. For instance, $t_{c}$ of audio and SLAM for the last event are $t_{c}=\nicefrac{0.1}{1.1}$ and $t_{c}=\nicefrac{1.0}{1.1}$, respectively.}
%   \vspace{0pt}
%   \label{fig:timelines}
% \end{figure}

%-------------------------------------------------------
\subsection{Planning Procedure}
\label{sec:technique:all}

The formal definition of context-aware task planning is described in Procedure~\ref{algo:task_planning}, the input of which is a list of inputs and sensors (\texttt{InputList}) and a configuration file (\texttt{ConfigFile}) that describes context-aware customization defined by the user and relative priorities of tasks. Initially, at lines~\ref{algo:line:input-start}--\ref{algo:line:input-end}, the system creates observable streams  for each input, and instantiates their scheduling score functions. At lines~\ref{algo:line:config1}--\ref{algo:line:config2}, by reading the configuration file, the system creates a task graph, similar to~\cref{sec:tasks} and \cref{sec:technique:controller}. The graph generation is out of the scope of this paper and we reuse the common methods from robotic development environments~\cite{quigley2009ros}. After initializing the scheduler at Line~\ref{algo:line:init_sch}, the processor only updates its setting based on the arrival of new events. On arrival of such an event that triggers the RX, the controller calculates a new scheduling score (Line~\ref{algo:line:score}) for that module, calculates a new dictionary of modulus and their scores at Line~\ref{algo:line:slice} based on Equations~\ref{equ:final_score_cfs} or~\ref{equ:final_score_rt}, and updates the time-slices in the OS scheduler at Line~\ref{algo:line:update_sch}.

%% file: tex/algo.tex
% \begin{table}[!b]
% \vspace{-0pt}
\setlength{\textfloatsep}{5pt}

\begin{algorithm}[t] 

  \SetAlgorithmName{Procedure}{}

  \scriptsize
  
  \SetKwInOut{Input}{Input}
  \SetKwInOut{Output}{Output}
  
  \SetKwFunction{place}{Place}
  \SetKwFunction{max}{Max}
  \SetKwFunction{min}{Min}
  \SetKwFunction{uplace}{UpdatePlace}
  
  \SetKwFunction{createStream}{CreateStream}
  \SetKwFunction{instantiateRX}{InstantiateRX}
  \SetKwFunction{createTaskGraph}{CreateTaskGraph}
  \SetKwFunction{addtolist}{AddToList}
  \SetKwFunction{initializeScheduling}{InitializeScheduling}
  \SetKwFunction{calculateScore}{CalculateScore}
  \SetKwFunction{calculateTimeSlices}{CalculateTimeSlices}
  \SetKwFunction{updateScheduler}{UpdateScheduler}
  
  \SetKwData{configfile}{ConfigFile}
  
  \SetKwArray{inputlist}{InputList}
  \SetKwArray{streamList}{StreamList}
  \SetKwArray{taskGraphList}{TaskGraphList}
  \SetKwArray{dictTimeSlices}{DictTimeSlices}
  
  \SetKw{KwGoTo}{go to}
  \SetKw{and}{and}
  
  \SetKwProg{myproc}{Event-Based Procedure}{}{}
  \SetKwProg{initial}{Initial}{}{}
  \SetKwProg{onStreamEvent}{OnObservableStreamEvent}{}{}
  
  \Input{\texttt{\configfile}: Configuration File with Context-Aware Setting \\
        and Relative Priorities of Tasks.}
  \Input{\texttt{\inputlist}: Input \& Sensor List}
  \initial{} {
    
    \For{input $\in$ \inputlist} {
    \label{algo:line:input-start}
    
        \tcp{Create observable stream.}
        stream $\gets$ \createStream{input}\;
        \tcp{Instantiate scheduling score function.}
        \instantiateRX{stream}\;
        \addtolist{stream, \streamList}\;
         \label{algo:line:input-end}
    }
    
    \For{context $\in$ \configfile} { \label{algo:line:config1}
        \tcp{Create context task graph.}
        graph $\gets$ \createTaskGraph{context}\;
        \tcp{Create combinator observable stream}
        contextStream $\gets$ \createStream{graph}\;
        \addtolist{contextStream, \taskGraphList} \label{algo:line:config2} \;
    }
    
    \tcp{Initialize an initial scheduling policy.}
    \initializeScheduling \label{algo:line:init_sch}\;
    
    \Return \streamList \taskGraphList
  }

  \tcp{On receiving an event on any stream after 
        its bound reactive function.}    
  \myproc{} {
    
    \onStreamEvent{stream}{
        \tcp{Calculate scheduling score for the stream.}
        \calculateScore{stream} \label{algo:line:score}\;
        \tcp{Calculate new real-time time-slices values.}
        \dictTimeSlices $\gets$ \calculateTimeSlices{} \label{algo:line:slice}\;
        \tcp{Update the scheduler.}
        \updateScheduler{\dictTimeSlices}\label{algo:line:update_sch}\;
    }
  }
  \caption{{\small Context-Aware Task Planning.}}
  \label{algo:task_planning}
\end{algorithm}

% \end{table}
% \vspace{-0pt}

%% file: tex/tesk-description.tex
This section provides implementation details of our specific tasks and their respective dataset.

% , and their accuracy/performance metrics.
%

\subsubsection{SLAM}
With a stereo camera input (Minoru3D~\cite{minoru}), we run the \texttt{ORB\_SLAM2}~\cite{murORB2} algorithm to localize the robot within its local environment. We use the EuRoC MAV dataset~\cite{burri2016euroc}. In addition to providing a stereo video input and ground truth values for error calculation, the EuRoC MAV dataset provides IMU sensor readings with accelerometer and gyroscope readings. This sensor data is used in deciding the scheduling of the SLAM module. Because of the computation demand of IMU, we implement its calculation on a separate module. These readings are fed into the controller, which creates an observable stream for each. Other modules (\eg, SLAM) then subscribe to these observable streams.

% The performance metric for this module is the average time taken to process a single pair of frames from the stereo camera. The accuracy metrics for this module are absolute trajectory error (ATE) and relative pose error (RPE) values.

% to integrate an IMU sensor on the Raspberry Pi to read accelerometer and gyroscope data.

% to output its desired prioritization value.

\subsubsection{Sign Detection}
The robot processes the images from its side cameras and uses a pre-trained neural network (trained on Street View House Numbers (SVHN) dataset~\cite{netzer2011reading}) to decide the room/street number for the signs. For experiments, the sign detection module is using the SVHN dataset test inputs.

% Figure~\ref{fig:room_number} shows an example of camera input and the desired output.

% The accuracy metric for this module is the error rate of the test data. The performance metric is the average time taken to process a single image.

%
% \begin{figure}[b]
%   \vspace{-15pt}
%   \centering
%   \includegraphics[width=1.0\linewidth]{fig/room-number.pdf}
%   \vspace{-15pt}
%   \caption{Room number detection: camera input and output.}
%   \vspace{-0pt}
%   \label{fig:room_number}
% \end{figure}
%

\subsubsection{Speech Detection}
The robot processes the microphone input from a microphone and uses the CMU Sphinx library (specifically, CMU PocketSphinx framework~\cite{huggins2006pocketsphinx}) for keyword spotting and later a DNN-based implementation~\cite{mozilladeep} to convert the speech to text. The speech detection module is using the Speech Commands dataset~\cite{warden2018speech}. This dataset includes a labeled set of various spoken commands.

% The accuracy metric for this module is the classification error rate of the input command. The performance metric for this module is the average time taken to process a command.

% \subsubsection{Inertial Measurement Unit (IMU)}

% \subsubsection{Chatbot}
% The robot gives feedback to users by processing the speech and generating appropriate responses by using a common chatbot tool, Chatterbot~\cite{chatbot}, which uses a selection of machine learning algorithms to produce a series of responses. After the response is generated, we use a library that uses OS text-to-speech engine to produce voice~\cite{pyttsx3}.

% \subsubsection{LIDAR}

\subsubsection{Navigation \& Arm Control}
For navigation, we send commands in the format specified in iRobot Create 2 Open Interface~\cite{irobotoi} through a serial port on iRobot. We also read several sensors and battery condition using this serial port. The navigation commands set the speed of each wheel separately.  Besides, for obstacle detection, we use a low-cost LIDAR sensor (360$\degree$ laser range scanner~\cite{lidar}). The LIDAR provides 360-degree scan field, 5.5hz/10hz rotating frequency with 8-meter ranger distance. We build a simple robot arm that works with Raspberry Pi on top of our robot~\cite{mearm}. The arm has simple grips and four servos to control. The module sends control commands to the arm to move and grab.

% Additionally, to mimic the computation~\cite{lidarslam}, we utilize a Lidar-based SLAM implementation, which is highly optimized to run as fast as C++-based implementations. 

% Thus, the module sends control commands for setting the speed to the iRobot.

% \subsubsection{Arm Control}

%% file: tex/res.tex
\setlength{\textfloatsep}{1\baselineskip}

We use iRobot Roomba~\cite{irobot} as our base navigation robot (Figure~\ref{fig:irobot}). We equip the robot with one Raspberry Pi 4~\cite{pi4}. The power source of the Pi is derived from the battery of iRobot with a voltage converter. The computation platform for all the modules is the Raspberry Pi.

% \renewcommand{\arraystretch}{0.9}
% \begin{table}[b]
%     \footnotesize
% 	\centering
% 	\vspace{-12pt}
% 	\caption{\small The specification of Raspberry Pi 4~\cite{pi4}.}
% 	\vspace{-5pt}
% 	\begin{tabular}{c | c | c}
% 		\toprule
%         CPU & \multicolumn{2}{c}{1.5\,GHz Quad Core ARM Cortex-72} \\
%         Memory &  \multicolumn{2}{c}{4\,GB LPDDR4-3200} \\
%         GPU & \multicolumn{2}{c}{No GPGPU Capability} \\
% 		\bottomrule
% 	\end{tabular}
% 	\label{tab:pi}
% 	\vspace{2pt}
% \end{table} \renewcommand{\arraystretch}{1}

%--------------------------------------------------------------
\subsection{Experiment Design \& Reproducibility }
\label{sec:res:design}

In our experiments, each task executes a pre-labeled dataset while we measure its performance as the controller adjusts the scheduling parameters dynamically. We use default Linux scheduler (CFS) as the baseline, and run context-aware (CA) configuration with two CFS and RT Linux schedulers (\cref{sec:technique:policy}), CFS CA and RT CA. To perform a fair comparison with the same set of experiments, we build an instrumentation tool, which uses a set of JSON files as timelines to artificially feed dataset inputs at certain times. In this way, we can execute the same set of experiments repeatedly with different schedulers. The timeline files are collected and constructed from a set of experiments from the measurements of the real robot. Each timeline file contains a series of inputs (\eg, images for SLAM and audio for speech recognition, or arm control commands) and their respective ground truth values (if any). We use these files to feed events to the controller while it calculates the parameters for the scheduler. We design experiments with different granularity, \emph{which includes all the implemented tasks in \cref{sec:tasks}}. Our first experiment, \texttt{exp1}, is the longest experiment with a total of three minutes footage, while \texttt{exp2} and \texttt{exp3} experiments are with shorter duration, one minutes and 15 seconds, respectively.

% Our instrumentation also ensures per-task accuracy/performance metrics are within acceptable ranges.

% iRobot executes all the tasks introduced in Section~\ref{sec:task-description} based on a context-aware configuration.
%-------------------------------------------------------
\subsection{Experimental Results}
\label{sec:res:exp}

\subsubsection{Proof of Concept}
\label{sec:res:exp:case}

\begin{figure}[t]
  \vspace{0pt}
  \centering
  \includegraphics[width=0.99\linewidth]{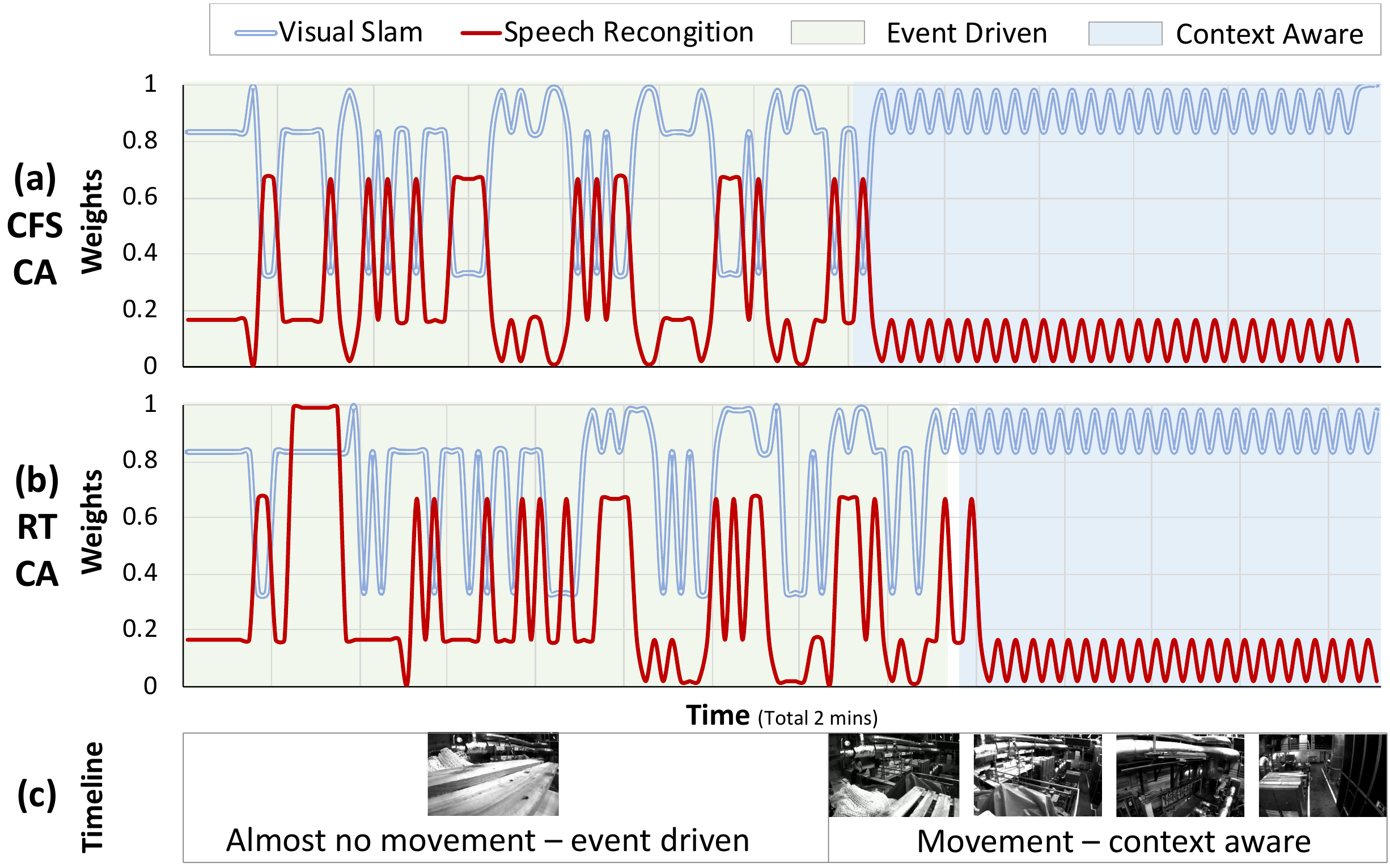}
  \vspace{-10pt}
  \caption{\small Normalized scheduling scores (weights) in a simple experiment with SLAM and speech recognition, (a) with CFS scheduler, and (b) with RT scheduler. Colored regions show the two phases of event-driven and context-aware philosophies with a timeline of sample frames in the footage (c).}
  \vspace{-10pt}
  \label{fig:weights-simple}
\end{figure}

To understand the intuition behind context-aware task handling, we present a proof-of-concept experiment. Figure~\ref{fig:weights-simple}a illustrates normalized scheduling scores (weights) with CFS scheduler for two main tasks, SLAM and speech recognition (and more sub-tasks such as camera and microphone inputs). The timeline, shown in Figure~\ref{fig:weights-simple}c, includes example footage from EuRoC dataset with addition of speech sounds, the beginning half of which has no movement. The context configuration by user prioritizes SLAM computation over speech recognition. When there is no determined context in the beginning half of the timeline, the weights are determined by the event-driven design as shown in Figure~\ref{fig:weights-simple}a. For instance, with a slight movement or upon a speech input, the weight of SLAM or speech recognition change accordingly. On the other hand, when robot moves, the weight for speech recognition is set to small values because of the \textit{context-aware} design and the user configurations that does not allow speech recognition while robot moves. Additionally, Figure~\ref{fig:weights-simple}b shows the normalized weights with RT scheduler. As seen, this scheduler has more lag in responding to changes since the scheduler uses a round-robin policy with dedicated time-slices per task. Figure~\ref{fig:weights-simple}c illustrates some frames from the timeline. In \cref{sec:res:exp:all_time}, we compare context-aware RT and CFS schedulers for all the experiments.

\subsubsection{Per-Task Accuracy Measurements}
\label{sec:res:exp:acc}

Throughout the three experiments, we observe accuracy changes as we changed scheduling configurations. However, in the baseline implementation with no context-aware task handling, where all tasks must run, accuracy drops in exchange for increased performance if the underlying computation modules are designed to sacrifice accuracy for real-time performance (\eg, SLAM dropping frames when computation takes longer than frame time). Thus, with context-aware scheduling, the resulted per-task accuracy are slightly higher.

%-------------------------------------------------------
\subsubsection{Per-Task Total Execution Time}
\label{sec:res:exp:time}

Our results show that the CFS CA parameter adjustments had the highest level of impact on per-task performance. For instance, speech recognition and sign detection tasks take a heavy performance penalty in \texttt{exp1} when the system  prioritized the SLAM module. Therefore, the configuration of the controller when running under the CFS CA is very important and must be carefully tuned. Generally speaking, the RT CA keeps a good balance between fair scheduling and using sensor events to prioritize the relevant modules at the right time. This is mainly because of  the technical restrictions with this scheduling policy, effectively making the RT CA policy an incremental improvement over the baseline CFS policy.

\begin{figure}[t]
  \vspace{0pt}
  \centering
  \includegraphics[width=0.99\linewidth]{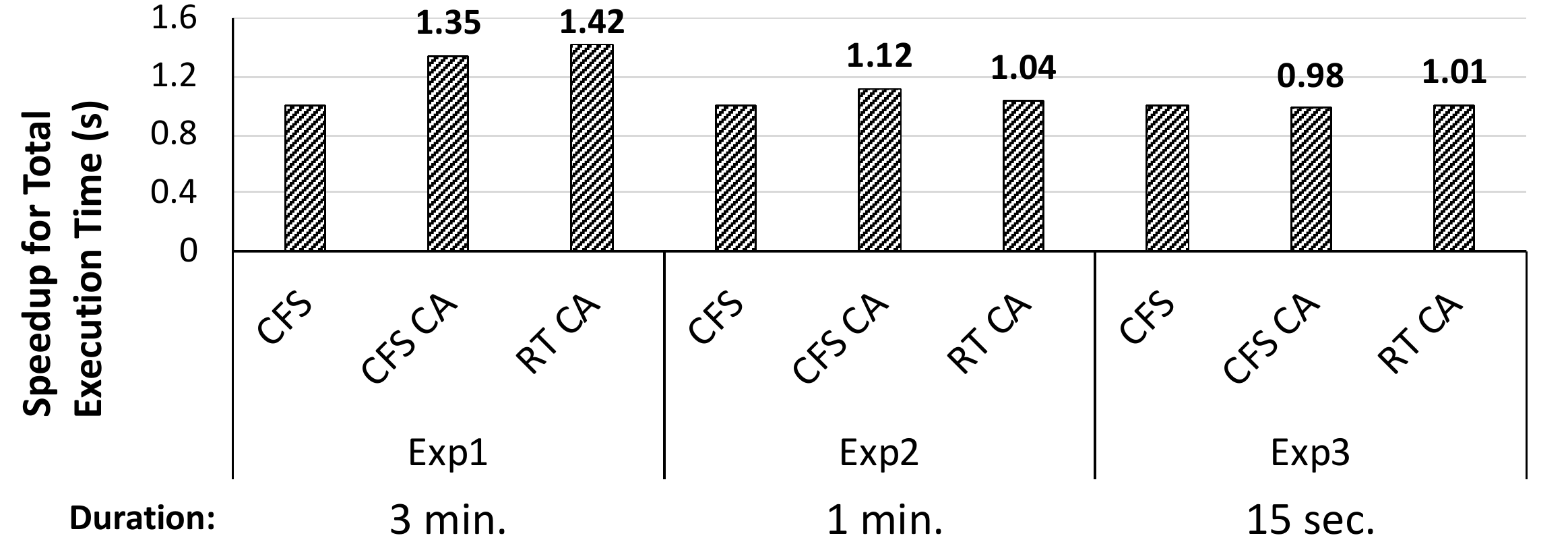}
  \vspace{-10pt}
  \caption{\small Speedup for total execution time with baseline, CFS context-aware (CFS CA), and RT contex-aware (RT CA).}
  \vspace{-10pt}
  \label{fig:total_time}
\end{figure}
\begin{figure}[b]
  \vspace{-0pt}
  \centering
  \includegraphics[width=0.99\linewidth]{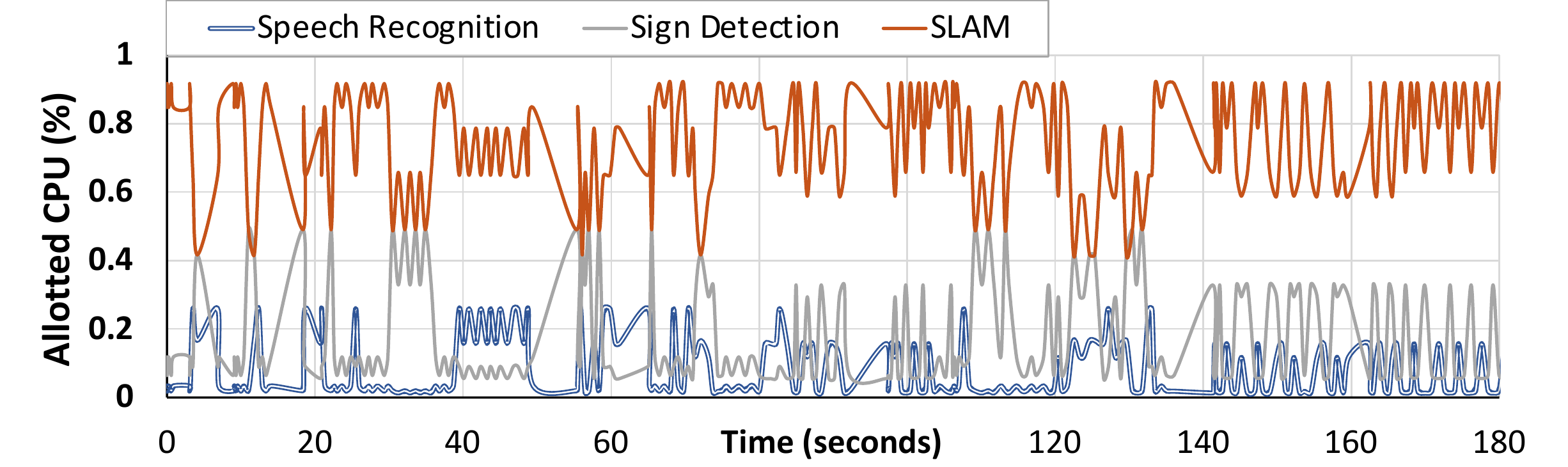}
  \vspace{-10pt}
  \caption{\small Allotted CPU shares for speech recognition, sign detection, and SLAM tasks during \texttt{exp1} using CFS CA.}
  \vspace{-0pt}
  \label{fig:weights}
\end{figure}
%

%-------------------------------------------------------
\subsubsection{Overall Execution Time}
\label{sec:res:exp:all_time}
Figure~\ref{fig:total_time} shows the speedup of different scheduler configurations for the total execution time of three experiments over the baseline scheduler, Linux CFS scheduler. As discussed in~\cref{sec:technique:policy}, our context-aware technique is implemented with two approaches, CFS and RT. As seen in Figure~\ref{fig:total_time}, with a longer run-time, our context-aware techniques  achieve up to 42\% speedup compared to the baseline, a significant speedup by only changing schedulers. As the run-time reduces, the context-aware configuration loses its impact and becomes less effective. Empirically, we found that higher run-time -- and therefore increased volumes of sensor data -- leads to higher speed-ups, compared to the baseline. Our experiments show this up to three minutes, but our experiments shows a similar trend with run-times beyond three minutes. This is because context-aware setting becomes more effective when there is a larger number of tasks within a longer execution time.

Figure~\ref{fig:weights} illustrates allotted CPU shares in percentage during the execution of \texttt{exp1}, the speedup of which is shown in Figure~\ref{fig:total_time}. This shows the underlying share per each task, which is directly related to the scheduling weights determined by the controller. The controller is using the context-aware CFS scheduler. As seen, since SLAM is more frequent and has more computations, most of the time the CPU is processing the SLAM task. However, when a relatively compute-intensive task requires more CPU (\eg, sign detection), more CPU shares are allocated to that task depending on the context. Meanwhile, as seen in the figure, sometimes a task allotted CPU share is zero, which is because of the context-aware configuration. As shown in Figure~\ref{fig:total_time}, the context-aware configuration allows us to achieve faster execution times.

% %
% \begin{figure}[h]
%   \vspace{-5pt}
%   \centering
%   \includegraphics[width=1.0\linewidth]{fig/slam-time}
%   \vspace{-15pt}
%   \caption{Average time per image for SLAM with and without context-aware policy.}
%   \vspace{-5pt}
%   \label{fig:slam_time}
% \end{figure}
% %

% %
% \begin{figure}[h]
%   \vspace{-5pt}
%   \centering
%   \includegraphics[width=1.0\linewidth]{fig/audio-time}
%   \vspace{-15pt}
%   \caption{.}
%   \vspace{-5pt}
%   \label{fig:audio_time}
% \end{figure}
% %

%% file: tex/conclusion.tex
In this paper, we introduced a context-aware task-handling for resource-constrained robots to extend their abilities with limited computation resources. We use reactive programming paradigm to build a lightweight controller that performs event-driven task scheduling using supported Linux kernel schedulers. Our system can dynamically schedule tasks in kernel-level by adjusting task scheduling parameters. We use containerized modules using Docker, which allows users to create and collaborate independently on several platforms. Finally, our experiments with Raspberry Pi 4 shows significant speedups while performing multiple tasks such as SLAM, sign detection, and speech recognition.

% Finally, we performed experiments on a modified iRobot Roomba 600s equipped with an Raspberry Pi 4 and ran various tasks such as SLAM, sign detection, and speech recognition. For the future work, we plan to include more tasks on new platforms to further test and improve our implementation. The main optimizations are improving the deterministic execution of Docker environments and integration with hardware resources. 